\documentclass[sigconf,nonacm]{acmart}
\AtBeginDocument{%
  \providecommand\BibTeX{{%
    \normalfont B\kern-0.5em{\scshape i\kern-0.25em b}\kern-0.8em\TeX}}}

\usepackage{multirow}
\usepackage{subcaption}
\usepackage{amsfonts,mathtools}

\usepackage{fixmath}

\begin{document}
\title{Absformer: Transformer-based Model for Unsupervised
Multi-Document Abstractive Summarization
}

\author{Mohamed	Trabelsi}
\email{mohamed.trabelsi@nokia-bell-labs.com}
\affiliation{%
 \institution{Nokia Bell Labs}
  \city{Murray Hill}
  \state{NJ}
  \country{USA}
}

\author{Huseyin Uzunalioglu}
\email{huseyin.uzunalioglu@nokia-bell-labs.com}
\affiliation{%
  \institution{Nokia Bell Labs}
  \city{Murray Hill}
  \state{NJ}
  \country{USA}
}

\begin{abstract}

Multi-document summarization (MDS) refers to the task of summarizing the text in multiple documents into a concise summary. The generated summary can save the time of reading many documents by providing the important content in the form of a few sentences. Abstractive MDS aims to generate a coherent and fluent summary for multiple documents using natural language generation techniques. In this paper, we consider the unsupervised abstractive MDS setting where there are only documents with no groundtruh summaries provided, and we propose Absformer, a new Transformer-based method for unsupervised abstractive summary generation. Our method consists of a first step where we pretrain a Transformer-based encoder using the masked language modeling (MLM) objective as the pretraining task in order to cluster the documents into semantically similar groups; and a second step where we train a Transformer-based decoder to generate abstractive summaries for the clusters of documents. To our knowledge, we are the first
to successfully incorporate a Transformer-based model to solve the unsupervised abstractive MDS task.  We
evaluate our approach using three real-world datasets from different domains, and we demonstrate both substantial improvements in terms of evaluation metrics over state-of-the-art abstractive-based methods, and generalization to datasets from different domains.
  
\end{abstract}




\maketitle

\section{Introduction}

In this era of big data, the fast increase of the availability of text data makes understanding and mining this data a very time-consuming task. Text summarization techniques help to analyze the text data efficiently and effectively. Important information can be summarized from multiple documents which refers to the case of multi-document summarization (MDS). Multiple deep learning (DL)-based methods for MDS rely on the presence of a large amount of annotated data in terms of document-summary pairs for supervised training. However, in many domains, such supervised data is not available and expensive to produce. Therefore, researchers have focused on the unsupervised learning setting where there are only documents available without gold summaries. Early works \cite{mei_www,titov_www,rossiello_eacl} in unsupervised MDS proposed extractive-based methods for unsupervised MDS. Then, researchers have focused on developing abstractive methods for MDS to generate more informative and concise summaries.

Existing methods in MDS incorporate recurrent architectures, such as LSTM and GRU, in autoencoder-based architectures. In recent years, Deep contextualized language models, like BERT \cite{Devlin2019BERTPO} and RoBERTa \cite{Liu2019RoBERTaAR}, have been proposed to solve multiple tasks \cite{dai2019,sakata2019,Chen2020TableSU,sun2019,khattab_sigir,nogueira_passage,nogueira_multi} by taking advantage of the Transformer \cite{transformer} architecture which has been shown to capture long-range dependencies better than the recurrent architectures. In particular, the Transformer-based seq2seq models have led to state-of-the-art results in supervised MDS. The Transformer-based models are not explored in the unsupervised MDS setting. For instance, incorporating Transformers in an autoencoder-based model for unsupervised MDS remains a challenging task because gold summaries are not available for training an end-to-end model. Therefore, the state-of-the-art end-to-end Transformer-based encoder-decoder models, such as PEGASUS \cite{pegasus}, cannot be directly applied to generate summaries.

We present \textit{\textbf{Absformer}}, a novel \textit{Trans\textbf{former}}-based model for unsupervised
multi-document \textit{\textbf{Abs}tractive} summarization. Our objective consists of predicting a summary, that is composed of a set of sentences, for each group of semantically similar documents. Therefore, Absformer is a two-phased method for unsupervised MDS. In phase I, we divide the set of documents into groups of semantically similar documents using either an unsupervised clustering algorithm when no criteria of partition is given, or a supervised classification when labeling information is available. We refer to this phase as the embedding-based document clustering which requires a pretrained or fine-tuned encoder in order to produce the embeddings of documents. We show that pretraining the encoder on a standard Masked Language Modeling (MLM) objective leads to clusters with semantically similar documents. In addition, these clusters can be enhanced when labeling information is available for fine-tuning of the encoder. In phase II, we generate a summary for every cluster using our new Transformer-based decoder. In the training phase, the decoder model generates the documents from the embeddings that are obtained from the frozen encoder. Then, the trained decoder is used to generate summaries from the embeddings of the centers of clusters as in previous works \cite{meansum,coavoux,brazinskas_acl}. An important design choice in Absformer is that the encoder should be kept frozen during the summary generation training in order to preserve the document clustering information in terms of document and cluster center embeddings. Therefore, compared to the traditional text-to-text encoder-decoder models, we decouple the encoder and decoder so that in phase II, Absformer takes directly an embedding-level information as input instead of the token-level information, and it generates as output a token-level sequence. We show that initializing the embedding layer, the decoder blocks, and the language model head of the Absformer's decoder using the corresponding weights of the frozen encoder leads to better results in terms of generated summaries and faster convergence in the training phase. We note that both the encoder and decoder has the same number of layers as DistilBERT which has a reduced size and comparable performance to BERT, because our objective is to obtain both accurate document clusters and concise summaries while keeping the time and memory complexity reasonable.

In summary, we make the following contributions: 
\begin{itemize}
    \item We propose a new Transformer-based method, called Absformer, for unsupervised MDS. Absformer is a two-phased method. In phase I, we propose to cluster the initial set of documents into groups of semantically similar documents by pretraining the encoder on a standard Masked Language Modeling (MLM) objective. The clusters can be enhanced when labeling information is available to fine-tune the encoder.
    \item In phase II, we train a Transformer-based decoder model that generates the documents from the embeddings that are obtained from the frozen encoder. Then, the trained decoder is used to generate summaries from the embeddings of the centers of clusters.
    \item  To improve the time and memory complexity of Absformer, we use the same architecture of DistilBERT for the encoder and decoder. To achieve faster convergence, we initialize the embedding layer, the decoder blocks with both self- and cross-attention, and the language model head of the Absformer's decoder using the corresponding weights of the frozen encoder.
    \item We experiment on three datasets (public and internal document corpus), and demonstrate that our new method outperforms the state-of-the-art abstractive-based baselines, and generalizes to document
collections from multiple domains.

\end{itemize}

\section{Related work}

\subsection{Multi-Document Summarization}

Multi-document summarization (MDS) \cite{survey_doc_sum} refers to the task of summarizing the text in multiple documents into a concise summary. Based on the availability of groundtruth summaries, MDS can be categorized into two groups: the first group is denoted as supervised MDS \cite{liu_acl,jin_acl,brazinskas_emnlp,li_acl,pasunuru_naacl,peter_iclr,zheng_emnlp,graph1,graph2,logan_emnlp,alexander_acl,ramakanth_2021,ziqiang_acl} when groundtruth summaries are available in the training phase, and unsupervised MDS \cite{rossiello_eacl,meansum,coavoux,nguyen_nlp,tampe_intellisys,brazinskas_acl} when there is no available groundtruth summaries in the training.

Existing summarization methods in MDS can be grouped into three groups namely: extractive, abstractive, and hybrid summarization. Extractive summarization techniques \cite{zheng_emnlp,cao_aaai,mao_emnlp,angelidis_emnlp,yasunaga_conll,wang_acl,extractive_rl,extractive_rl2,extractive_rl3,bert_extractive_summ} select salient parts from multiple documents in order to form a concise summary. The extractive methods are usually composed of two components: sentence ranking and sentence selection \cite{ziqiang_aaai,ramesh_aaai}. Abstractive MDS \cite{zhang_icnl,li_emnlp,wang_naacl,lebanoff_emnlp,fabbri_acl,bart,mass,sascha_acl,li_neurips,sergey_acl,sebastian_emnlp} aims to generate a coherent and fluent summary for multiple documents using natural language generation techniques. Hybrid MDS \cite{amplayo_eacl} aims to combine the extractive and abstractive summarization techniques in order to improve the quality of the generated summaries.

In the past few years, DL has led to a significant
improvement in multiple tasks, where DL-based methods achieved
state-of-the-art results for text, image, and speech data. DL models are trained end-to-end to automatically
extract features and build models. This significantly reduces the human effort that is needed in traditional methods for
feature engineering, and gives the model the ability to capture
specific features that are better than the hand-crafted ones for
multiple tasks. Following the success of DL models, researchers
have focused on exploring DL in MDS. Different neural architectures are used for MDS. For instance, recurrent architectures, including Recurrent Neural Networks (RNNs), Long Short-Term Memory (LSTM), and Gated Recurrent Unit (GRU), have been proposed as the main component in multiple methods \cite{meansum,coavoux,zhang_icnl,li_emnlp}. In addition, Convolutional neural networks (CNNs) have been also incorporated into multiple MDS methods \cite{oposum,ziqiang_aaai_2017}. The recent success of graph neural networks has boosted research on various tasks including MDS \cite{yasunaga_conll,wang_acl}.

In recent years, Deep contextualized language models, like BERT \cite{Devlin2019BERTPO}, RoBERTa \cite{Liu2019RoBERTaAR}, DeBERTa \cite{deberta}, and DistilBERT \cite{distilbert} have been proposed to solve multiple information retrieval and natural language processing (NLP) tasks \cite{strubert,dame,selab,dai2019,sakata2019,Chen2020TableSU,sun2019,khattab_sigir,nogueira_passage,nogueira_multi,survey_doc_retrieval,selab_arxiv}. 
 Different from traditional word embeddings, the pretrained neural language models are contextual where the representation of a token is a function of the entire sentence. This is mainly achieved by the use of a self-attention structure called a Transformer \cite{transformer}. 
Building on the Transformer architecture, multiple methods \cite{liu_acl,jin_acl,brazinskas_emnlp,li_acl,pasunuru_naacl,peter_iclr,openai1,openai2,openai3} achieved state-of-the-art results in supervised MDS.

\subsection{Unsupervised Multi-Document Summarization}

For unsupervised MDS, early works \cite{mei_www,titov_www,rossiello_eacl} introduced extractive-based methods for unsupervised MDS. Then, researchers have focused on developing abstractive methods for MDS. LSTM is used as the main component in multiple methods for the unsupervised MDS \cite{meansum,coavoux,amplayo_acl}. Chu et al. \cite{meansum} proposed an auto-encoder based method, called Meansum, where LSTM is incorporated in both the encoder and decoder. Meansum generates summaries for a fixed number of documents (8 documents). The summarization process in Meansum starts by averaging the embeddings of documents that are obtained from the LSTM-based encoder in order to compute the embedding of the summary. Then, the resulting embedding is forwarded to an LSTM-based decoder to generate the summary. The training of the encoder and decoder is monitored using two loss functions namely the average summary similarity loss and the auto-encoder reconstruction loss. The former loss constrains the generated summary to be semantically similar to the original documents, and the latter loss guides Meansum to reconstruct the original documents using a sequence-to-sequence cross-entropy loss with teacher forcing.

LSTM was also used as a main component in the method proposed by Coavoux et al. \cite{coavoux}. The authors proposed an unsupervised abstractive summarization neural model for opinion summarization from product reviews. This method uses an aspect-based clustering technique that groups reviews that are about the same aspect. Then, an LSTM-based decoder generates a summary for each aspect. The decoder is initialized by the cluster representation in order to generate a sequence of tokens as a summary. The cluster representation is computed using the average of the embeddings of the most salient sentences within each cluster. The saliency scores are determined using the clustering model that assigns a confidence score to each sentence in the reviews. For the summary decoding, top-k sampling decoding \cite{fan_acl} is used to predict the next token. 

Amplayo et al.\cite{amplayo_acl} proposed a method called DenoiseSum where denoising autoencoders is used for unsupervised summary generation. Multiple noisy versions of reviews are created using word-, sentence-, and document-level noising techniques. LSTM is used in both the encoder to obtain contextualized embeddings, and the decoder to denoise the reviews and generate the summaries.

Multiple unsupervised MDS methods \cite{nguyen_nlp,brazinskas_acl,hayate_emnlp,isonuma_ling} incorporate the variational autoencoder (VAE) model to generate summaries. Brazinskas et al. \cite{brazinskas_acl} proposed a VAE-based model, called CopyCat, for opinion summarization. CopyCat is trained using two types of loss functions: a reconstruction loss that recovers the original reviews from the latent representations, and a Kullback-Leibler (KL) divergence that penalizes the deviation of the computed posteriors from the priors. At the generation phase, the latent vectors of reviews are averaged to compute the summary representation that is decoded to generate the summary tokens. 

Iso et al. \cite{hayate_emnlp} showed that the averaging aggregation can cause the summary degeneration problem that leads to generic summaries. The authors proposed a method, called convex aggregation for opinion summarization (COOP), that computes a weighted average of the latent vectors of reviews as the summary representation, with the weights are estimated to maximize the word overlap between the generated summaries and the reviews. Nguyen et al. \cite{nguyen_nlp} proposed a class-specific VAE-based model for unsupervised review summarization. The model uses an independent classifier that assigns each review to the predefined classes. Isonuma et al. \cite{isonuma_ling} uses a recursive Gaussian mixture to model the sentence granularity in a latent space, and generate summaries with multiple granularities.

\section{Problem Statement}

In the unsupervised MDS, multiple documents $D= \{d_1,d_2,\ldots,d_n\}$ are given, where $n$ is the total number of documents. The task consists of predicting a summary, that is composed of a set of sentences, for each group of similar documents. This means that the set of documents $D$ should be divided into similar groups using either an unsupervised clustering algorithm when no criteria of partition is given, or a supervised classification when labeling information is available. For the rest of the paper, we refer to the step of grouping similar documents as the document clustering regardless of the technique that is used for this purpose. The document clustering step results in $K$ groups of documents $D =\{g_1,g_2,\ldots,g_K\}$, where every group $g_i$ contains similar documents (hard assignment is used in the document clustering step which means that a given document $d_j \in D$ can belong to only one group). The objective is to learn two models $F$ and $N$, where (1) $F$ is used for document clustering step in order to partition the input space, and (2) $N$ is used to generate a coherent summary for every cluster obtained in the document clustering step.

\section{Absformer: Document Clustering}

\begin{figure*}[ht!]
\centering
\includegraphics[width=0.80\textwidth]{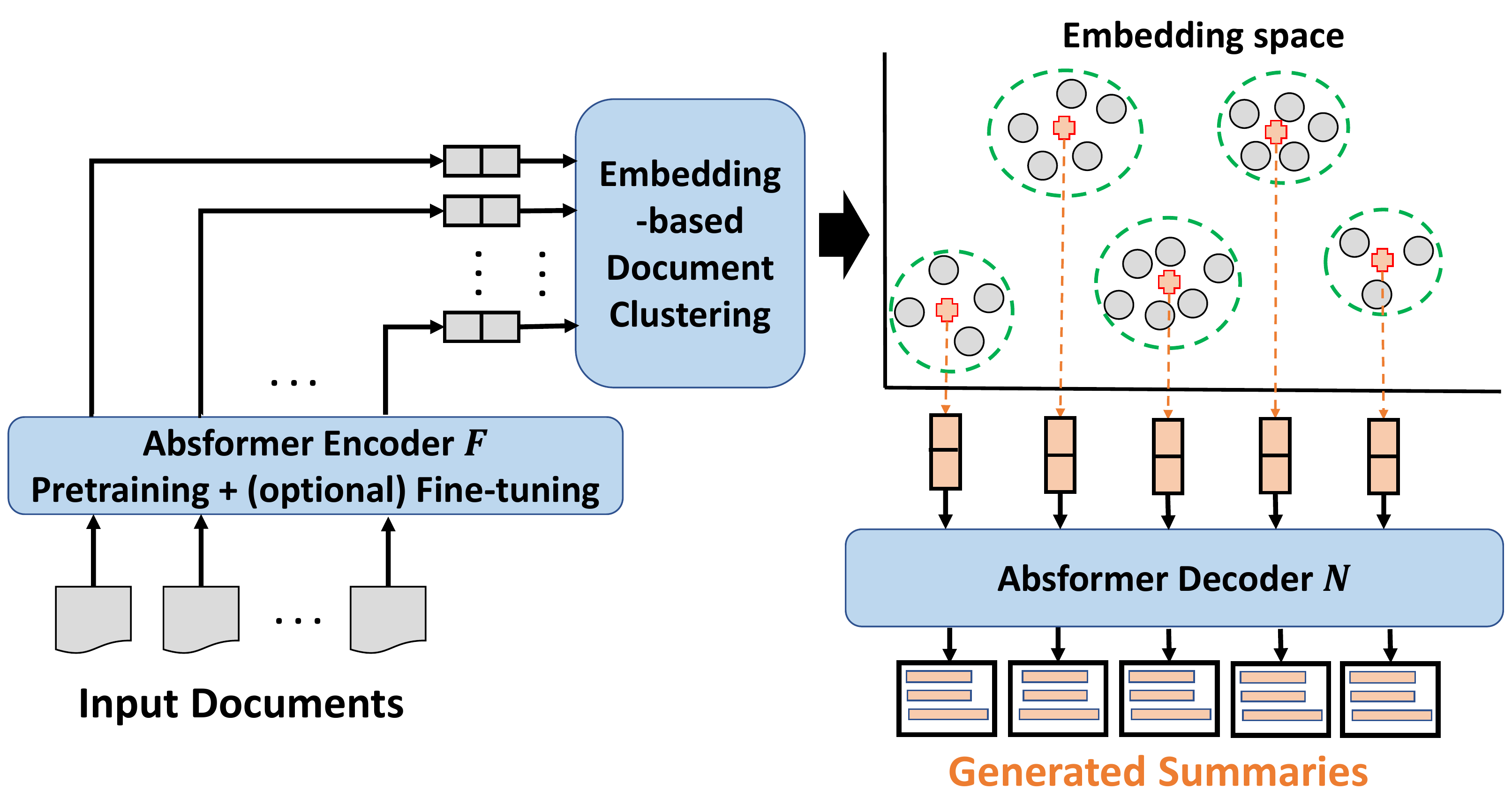}
\caption{Documents are clustered into semantically similar groups by pretraining the encoder $F$ on a standard Masked Language Modeling (MLM) with an optional fine-tuning when labeling information is available. Then, a trained decoder $N$ is used to generate summaries from the embeddings of the centers of clusters.}
\label{absformer_arch}
\end{figure*}

In this section, we introduce our proposed method Absformer which
is a two-phased Transformer-based method for abstractive MDS. We first describe phase I which consists of the document clustering step, and then present phase II which consists of the summary generation step. The overview of our method is shown in Figure \ref{absformer_arch}.

Existing methods in unsupervised MDS \cite{meansum,coavoux} use LSTM as the main component to obtain either document- or sentence-level representations. In our proposed model, we use the Transformer as the main component in both phases. In particular, we use DistilBERT, which is formed of multiple layers of Transformer blocks, to obtain document-level representations. We choose to use DistilBERT as the
main component for the clustering and generation models for two reasons. First, DistilBERT takes advantage of the Transformer’s self-attention which captures long-range dependencies better
than the recurrent architectures. Second,  DistilBERT has a reduced size and comparable performance to BERT, and our objective is to obtain both accurate document clusters and coherent summaries while keeping the time and memory complexity reasonable.

\subsection{Pretraining of Encoder}

We use a standard Masked Language Modeling (MLM) objective with a
masking rate of 15\% sub-tokens in a document. Formally, Given a sequence
of natural language tokens $\boldsymbol{x}=x_{1}, x_{2}, \ldots, x_{n}$, we train a DistilBERT masked language model that aims to recover the sub-tokens in $\boldsymbol{x}$ from the masked sequence. The mask special token, denoted by $[\text{MASK}]$, is used to replace a subset of tokens $\boldsymbol{x}_{m}=\left\{x_{i_{1}}, \ldots, x_{i_{m}}\right\}$ in $\boldsymbol{x}$ in order to obtain the masked sequence $\widetilde{\boldsymbol{x}}$. A masked DistilBERT language model computes the probability distribution $p_{\boldsymbol{\theta}}\left(\boldsymbol{x}_{m} \mid \tilde{\boldsymbol{x}}\right)$ over the masked tokens, with $\boldsymbol{\theta}$ are the parameters of the masked DistilBERT language model.

The parameters $\boldsymbol{\theta}$ are updated by maximizing the $p_{\boldsymbol{\theta}}\left(\boldsymbol{x}_{m} \mid \tilde{\boldsymbol{x}}\right)$ using the initial collection of documents $D$. After the pretraining phase, the pretrained DistilBERT is used as an encoder $F$ to compute representations for documents, and the parameters $\boldsymbol{\theta}$ are either jointly tuned with a multi-layer perceptron (MLP) if label information is used during the embedding-based document clustering, or kept frozen if unsupervised clustering is used on top of the computed document representations.

\subsection{Embedding-based Document Clustering}

We generate a summary for each group of documents that are semantically similar. The pretrained DistilBERT model $F$ captures semantic matching signals by using the MLM objective as the pretraining task. Therefore, the document embeddings that are computed by $F$ can be used as input to an unsupervised clustering algorithm to obtain clusters of semantically similar documents. The representation of a document $d_i$, denoted by $Rep(d_i)$, is given by:

\begin{equation}
Rep(d_i)=[\text{CLS}] x_{1} x_{2} \ldots x_{n_i} [\text{SEP}]
\end{equation}
where $d_i = x_{1} x_{2} \ldots x_{n_i}$ is a sequence of $n_i$ tokens,  and $[\text{CLS}]$ and $[\text{SEP}]$ are DistilBERT special tokens that are added into
the sequence similar to the single sentence classification setting. 

The encoder $F$ takes as input the representation of a document $d_i$, denoted by $Rep(d_i)$, in order to compute the embedding $\overline{\boldsymbol{d_i}} \in \mathbb{R}^{h}$ that is extracted using the
hidden state of the $[\text{CLS}]$ token from the last Transformer block in the DistilBERT model $F$, where $h$ is the dimension of the embedding.

\subsubsection{Cluster Documents Without Label Information} The Transformer based document embeddings $\overline{\boldsymbol{D}} = \{\overline{\boldsymbol{d_1}}, \overline{\boldsymbol{d_2}}, \ldots, \overline{\boldsymbol{d_n}}\}$ are forwarded to an unsupervised clustering step to obtain $K$ clusters of documents. In this paper, we use the k-means clustering algorithm which has been shown to be both effective in terms of clustering results and efficient in terms of time and memory complexity. We leave investigating more advanced unsupervised clustering algorithms as a future direction.

As a result of the clustering step, we obtain $K$ clusters of documents $D =\{g_1,g_2,\ldots,g_K\}$, where every cluster $g_i = \{d^i_1,d^i_2, \ldots, d^i_{|g_i|}\}$ is composed of $|g_i|$ documents from $D$ and a cluster center $C_{g_i}$. We denote the cluster embedding that is obtained from the k-means step by $\overline{\boldsymbol{C_{g_i}}}$. To reduce the outliers effect, we compute the cluster-level weight of every document which is given by:

\begin{equation}
w_{d^i_j}=\frac{\min _{d^i_k \in g_{i}} dist(C_{g_i},d^i_k)}{dist(C_{g_i},d^i_j)}
\end{equation}
where:
\begin{equation}
dist(C_{g_i},d^i_k) = L^2_{norm}(\overline{\boldsymbol{C_{g_i}}},\overline{\boldsymbol{d^i_k}})
\end{equation}
$L^2_{norm}(v_1,v_2)$ denotes the Euclidean norm between two vectors $v_1$ and $v_2$. Then, we update the center embedding $\overline{\boldsymbol{C_{g_i}}}$ of every cluster $g_i$ using the cluster-level weights of documents:

\begin{equation}
\label{compute_center}
\overline{\boldsymbol{C_{g_i}}}=\frac{\sum_{j=1}^{|g_i|} w_{d^i_j} \overline{\boldsymbol{d^i_j}}}{\sum_{j=1}^{|g_i|} w_{d^i_j}}
\end{equation}
The embeddings of documents and centers that are obtained from the encoder $F$ are forwarded to the summary generation step, and kept frozen during both summary generation training and inference.

\subsubsection{Cluster Documents With Label Information} The unsupervised clustering step leads to semantically similar clusters as a result of the MLM objective. There are cases where summaries should be generated for documents that share more common characteristics in addition to the semantic similarity. For example, if $D$ is composed of product reviews, an interesting task is to generate a summary for similar products or similar ratings. In other words, the labeling information or the clustering criteria in this case is explicitly provided either by the product type or the rating score. Therefore, the pretrained DistilBERT model should be fine-tuned using the available labeling information to obtain semantically similar clusters with common characteristics. 

In the fine-tuning phase, each document $d_i \in D$ is associated with a label $l_i \in L$, with $L$ is the set of labels (e.g. rating scores, and product types). The pretrained DistilBERT model is used as an encoder $F$, and its parameters $\boldsymbol{\theta}$ are jointly trained with a task-specific softmax layer $W$ that is added on top of DistilBERT to predict the probability of a given label $l \in L$:
\begin{equation}
p\left(l \mid \overline{\boldsymbol{d_i}}^{\boldsymbol{\theta}}\right)=\operatorname{softmax}\left(W \overline{\boldsymbol{d_i}}^{\boldsymbol{\theta}}\right)
\end{equation}
The parameters of DistilBERT, denoted by $\boldsymbol{\theta}$, and the softmax layer parameters $W$  are fine-tuned by maximizing the log-probability of the true label.

The fine-tuning phase results in $|L|$ clusters, with $|L|$ is the number of labels of $L$. Similar to clustering documents without label information, each cluster should be associated with a cluster center, and each document should belong to one cluster and have a membership weight to reduce the effect of outliers. For a given document $d_i$, the first step is to compute the probability distribution of $d_i$ over all the labels from $L$:
\begin{equation}
p_i=\operatorname{softmax}\left(W \overline{\boldsymbol{d_i}}^{\boldsymbol{\theta}}\right)
\end{equation}
where $p_i \in \mathbb{R}^{|L|}$ is the vector of predicted probabilities. The second step is to predict both the cluster of $d_i$, denoted by $c_i$, and the membership weight, denoted by $w_i$:

\begin{equation}
\begin{array}{l}
c_i = \underset{l \in L}{\operatorname{argmax}} \text{ } p_i[l] \\
w_i = \underset{l \in L}{\operatorname{max}} \text{ } p_i[l]
\end{array}
\end{equation}
The cluster centers are computed similar to the case of clustering documents without label information using Equation (\ref{compute_center}). Finally, we obtain $|L|$ clusters of documents $D =\{g_1,g_2,\ldots,g_{|L|}\}$, where every cluster $g_i = \{d^i_1,d^i_2, \ldots, d^i_{|g_i|}\}$ is composed of $|g_i|$ documents from $D$ and a cluster center $C_{g_i}$.

For the rest of the paper, we denote the number of clusters by $K$ regardless of the presence or absence of the label information during the clustering phase. For example, Figure \ref{absformer_arch} depicts the case where $K=5$. In conclusion, the document clustering phase results in (1) semantically similar clusters $\{g_1,g_2,\ldots,g_K\}$  characterized by cluster centers $\overline{\boldsymbol{C_{g_1}}}, \overline{\boldsymbol{C_{g_2}}}, \ldots, \overline{\boldsymbol{C_{g_K}}}$, respectively, and documents with membership weights $w_{d^{c(i)}_i}$, where $c(i)$ is the cluster associated with $d_i$, and (2) trained DistilBERT-based encoder model $F$ that is used to map documents into the embedding space. The outputs of the document clustering phase are forwarded to the summary generation step in order to generate a coherent abstractive summary for every cluster $g_k$.

\section{Absformer: Summary Generation}

Pretrained Transformer-based models have led to state-of-the art results in text summarization. For instance, PEGASUS \cite{pegasus} is an encoder-decoder Transformer-based model that is used for supervised abstractive summarization. The groundtruth summaries are given in supervised abstractive summarization, so that both the encoder and decoder are jointly trained. In this paper, we address the unsupervised abstractive summarization so that end-to-end Transformer-based encoder-decoder models cannot be directly applied to generate summaries. We propose to train a Transformer-based decoder model $N$ that generates the documents from the embeddings that are obtained from the frozen encoder $F$. Then, the trained decoder $N$ is used to generate summaries from the embeddings of cluster's centers as shown in Figure \ref{absformer_arch}.

\subsection{Encoder-Decoder Model}

An important design choice in Absformer is that the encoder $F$ should be kept frozen during the summary generation training in order to preserve the document clustering information in terms of document and cluster center embeddings as shown in the embedding space of Figure \ref{absformer_arch}. Therefore, we directly used the embeddings $\{\overline{\boldsymbol{d_1}}, \overline{\boldsymbol{d_2}}, \ldots, \overline{\boldsymbol{d_n}}\}$ as input to the decoder model. Compared to the traditional text-to-text encoder-decoder models, the decoder part of Absformer takes an embedding-level information as input instead of the token-level information, and it generates as output a token-level sequence.

Our proposed decoder has the same number of blocks as the DistilBERT-based encoder model $F$. Each block in $F$ is composed of a self-attention layer and feed forward layer. To incorporate the document embedding into the decoding process, we also include a cross-attention layer in the decoder of Absformer similar to T5 model \cite{t5}. In addition to the decoder blocks, the DistilBERT-based decoder model is composed of an embedding layer and a language model head.

\subsubsection{Embedding Layer} In each timestamp, the decoder generates a token in order to either generate the document tokens from the input embedding, or the summary tokens from the embedding of the cluster's center. The token-level embedding is obtained from an embedding layer in the decoder $N$. This embedding layer is composed of word embeddings and position embeddings, and is initialized using the encoder's embedding layer to incorporate the token-level knowledge captured by the encoder $F$. The output of the embedding layer is forwarded to the first decoder block of $N$.

\subsubsection{Decoder Block} Each block is composed of three components which are the self-attention head, cross-attention head, and feed forward layer. 

\textbf{Self-attention Head:} This is a Transformer-based layer with 12 attention heads. There are 6 self-attention layers in total, and each layer is initialized using the respective self-attention from the frozen encoder $F$. The initialization from the encoder $F$ leads to faster convergence and lower loss than randomly initializing the self-attention head. The Transformer-based self-attention heads capture long-range dependencies better
than the recurrent architectures, and this leads to more accurate generated tokens even for long sequences (maximum number of tokens is 512). The deep contextualized embeddings, that are obtained at the timestamp $t$, are forwarded to the cross-attention head.

\textbf{Cross-attention Head:} The embeddings that are obtained from the self-attention head depends only on the generated tokens up to the timestamp $t$. To incorporate the encoder embeddings into the decoding process, we add a cross-attention head to compute both context- and encoder-aware embeddings. The cross-attention is also a Transformer-based layer with 12 attention heads. There are 6 cross-attention layers in total.

Formally, in a given attention head from a cross-attention layer, three parametric matrices are introduced: a query matrix $ Q\in \mathbb{R}^{h \times h}$, a key matrix $K \in \mathbb{R}^{h \times h}$, and a value matrix $V \in \mathbb{R}^{h \times h}$. For a given document $d_i$, suppose that the embedding of the $t$ generated tokens is denoted by $E_t \in \mathbb{R}^{t \times h_a}$, the cross-attention between the token embeddings $E_t$ and the encoder embedding $\overline{\boldsymbol{d_i}}$ is given by:

\begin{equation}
\begin{array}{l}
\mathcal{Q}=E_t Q \in \mathbb{R}^{t \times h} \\
\mathcal{K}=\overline{\boldsymbol{d_i}} K \in  \mathbb{R}^{1 \times h}  \\
\mathcal{V}=\overline{\boldsymbol{d_i}} V \in  \mathbb{R}^{1 \times h} \\
Att(E_t,\overline{\boldsymbol{d_i}}, Q, K, V)=\operatorname{softmax}\left(\frac{\mathcal{Q} \mathcal{K}^{T}}{\sqrt{h}}\right) \mathcal{V} \in \mathbb{R}^{t \times h}
\end{array}
\end{equation}
So, the key output $\mathcal{K}$ and value output $\mathcal{V}$ are computed using the encoder embedding $\overline{\boldsymbol{d_i}}$, and the query output is computed using the generated tokens embeddings $E_t$. The context- and encoder-aware embeddings that are obtained from the cross-attention head are forwarded to the feed forward layer. We note that we also initialized the cross-attention heads using the self-attention parameters from the frozen encoder $F$.

\textbf{Feed Forward Layer:} This layer is composed of a feed forward network with 2 linear layers, and a normalization layer. These neural components are initialized using the corresponding feed forward layer of the encoder $F$. The output of the feed forward layer of a block $l$ is forwarded to the self-attention head of the block $l+1$.

\subsubsection{Language Model Head} The next token prediction is used to train the decoder $N$. In other words, in each timestamp $t$, the decoder $N$ should output the token that corresponds to the position $t+1$. The language model head takes as input the embedding of the sequence of length $t$ that is obtained from the last decoder block, and outputs a probability distribution over all the vocabulary of the decoder $N$ (same vocabulary as the encoder $F$). For faster convergence of the decoding training, we initialize the language model head of the  decoder using the language model head of pretraining the encoder $F$ on the MLM in the document clustering phase. The language model head of the decoder is composed of a first linear layer, GELU activation, normalization layer, and a second linear linear with a dimension of output layer that is equal to the size of the decoder's vocabulary.

\subsection{Decoder Training}

The decoder $N$ is trained using the next token prediction task with teacher forcing. Instead of using a standard
cross-entropy loss to reconstruct the original documents, we train our decoder $N$ using a weighted cross-entropy loss in order to focus the efforts of training on reconstructing documents that are close to the cluster's center. Therefore, we reduce the effect of outliers on the generated summary. Formally, the decoder loss function $L_N$ of a batch of documents of size $B$ is given by:

\begin{equation}
L_{N}=-\sum_{i \in B} \sum_{j=1}^{T_i} w_{d^{c(i)}_i} \log P\left(t_j \mid t_0^{j-1}\right)
\end{equation}

Where $T_i$ is the length of document $d_i$, $t_j$ is the $j$-th token of the document $d_i$, and $t_0^{j-1}$ is the sequence of document $d_i$ up to token $j-1$.

\subsection{Summary Decoding}

After training the decoder $N$ on reconstructing the documents from the embeddings that are obtained from the encoder $F$, the last step is to generate a summary for each cluster. Each cluster's center $\overline{\boldsymbol{C_{g_i}}}$ is used as input to the decoder $N$ as shown in Figure \ref{absformer_arch}. More specifically, the cross-attention head of each decoder block uses the cluster's center embedding $\overline{\boldsymbol{C_{g_i}}}$ to compute the key and value matrices, so that we obtain a cluster-aware embedding for each timestamp as output from the last decoder block. The output embedding in each timestamp is forwarded to the language model head of the decoder $N$ in order to compute the probability distribution over all the vocabulary.

Multiple methods \cite{fan_acl,Holtzman2020The} are proposed in the literature to select the next token from the computed probability distribution over the vocabulary. We experimentally found that a combination of top-p \cite{Holtzman2020The} and top-K \cite{fan_acl} samplings lead to the best quality of generated summaries. In particular, (1) with top-p sampling, it is possible to dynamically increase or decrease the set of candidate tokens according to the probability distribution of the next word, and (2) with top-K sampling, the tokens with very low probability distribution can be discarded. We perform the summary decoding $S$ times so that we obtain $S$ different summaries for each cluster.

Similar to Coavoux et al. \cite{coavoux}, the $S$ generated summaries for a given cluster $g_i$ are ranked based on the cosine similarity between the embedding of each summary and the cluster's center embedding $\overline{\boldsymbol{C_{g_i}}}$ in order to keep only the semantically similar summaries to each of the clusters. The embeddings of summaries are computed using the encoder $F$.

\begin{table*}[t!]
\begin{subtable}[t]{0.48\textwidth}
\centering
\begin{tabular}{@{}llll@{}}
\toprule
\bf Method Name & R-1 & R-2 & R-L   \\ \midrule
 Aspect + MTL \cite{coavoux}& 30.00   & 5.00  & 17.00   \\
 MeanSum \cite{meansum}& 30.16  & 4.51  & 17.76  \\
 Copycat \cite{brazinskas_acl}& 31.84 & 5.79  & \textbf{20.00}  \\
 DenoiseSum \cite{amplayo_acl}& 34.82  & 6.12 & 18.58\\
 RecurSum \cite{isonuma_ling}& 34.91   & 6.33  & 18.91 \\\bottomrule
Absformer (w/o labeling information)   & 34.96   & 7.05 & 18.85 \\
Absformer (w/o decoder's initialization)   & 35.85   & 7.12 & 19.14  \\
Absformer (w/o pretraining)   & 35.52  & 7.08 & 19.11   \\
Absformer (unweighted cross-entropy)   & 36.34 & 7.20 & 19.22  \\
Absformer& \textbf{37.76}  & \textbf{8.73}  & \textbf{20.00} 
 \\ \bottomrule
\end{tabular}
\caption{\footnotesize Amazon}
\label{tab:table1_d}
\end{subtable}
\hspace{\fill}
\begin{subtable}[t]{0.45\textwidth}
\centering
\begin{tabular}{@{}llll@{}}
\toprule
\bf Method Name & R-1 & R-2 & R-L   \\ \midrule
 Aspect + MTL \cite{coavoux}& 28.12  & 3.68   &  15.23  \\
 MeanSum \cite{meansum}& 28.66  & 3.73  & 15.77  \\
 Copycat \cite{brazinskas_acl}& 28.95 & 4.80  & 17.76  \\
 DenoiseSum \cite{amplayo_acl}& 29.77  & 5.02 & 17.63\\
 RecurSum \cite{isonuma_ling}& 33.24   & 5.15  & \textbf{18.01} \\\bottomrule
Absformer (w/o labeling information)   & 33.75  & 6.12 & 16.23 \\
Absformer (w/o decoder's initialization)   & 34.86  & 6.82 & 16.88  \\
Absformer (w/o pretraining)   & 34.14  & 6.25 & 16.56   \\
Absformer (unweighted cross-entropy)   & 35.92 & 7.12 & 17.36  \\
Absformer& \textbf{36.87}  & \textbf{8.05}  & 17.87 
 \\ \bottomrule
\end{tabular}
\caption{\footnotesize Yelp}
\label{tab:table1_a}
\end{subtable}
\caption{ ROUGE scores on publicly available benchmarks.}
\label{metrics}
\end{table*}

\section{Evaluation} \label{eval}

\subsection{Data Collections}

\subsubsection{Amazon Reviews} This dataset is composed of the Oposum corpus \cite{oposum} which contains 3,461,603 Amazon reviews that are collected from 6 types of products (Bags and cases, Bluetooth, Boots, Keyboards, TV, Vacuums), and extracted from the Amazon Product Dataset \cite{amazon}. Each review is accompanied by a 5-star rating . For each product, there are 30 gold summaries that are used only to evaluate the generated summaries by our proposed model and baselines during the testing phase.

\subsubsection{Yelp Reviews} This dataset includes a large number of reviews without gold-standard summaries. Each review is accompanied by a 5-star rating. Similar to Chu et al. \cite{meansum}, Businesses with less than 50 reviews are removed, so that there are enough reviews to be summarized for each product. This dataset contains 1,297,880 reviews. We consider 7 types of products (Shopping, Home Services, Beauty \& Spas, Health \& Medical, Bars, Hotels \& Travel, Restaurants). For each product, there are 20 gold summaries that are used only to evaluate the generated summaries by our model and baselines during the testing phase.

\subsubsection{Ticket Data from Network Equipment} Our work is motivated by the business need to summarize ticket data from network equipment. When an issue occurs in mobile networks, a ticket is opened to describe the problem at various levels of detail. Each ticket contains multiple pieces of information describing the issue. The ticket is usually resolved by a unit that is usually determined by the tester. In order to accelerate the process of resolving tickets, the tester can take advantage of the previous tickets to resolve common problems. However, it is a very time consuming task for the tester to read the previous tickets in order to find similar issues. Therefore, summarizing groups of similar tickets into a concise summary can help the tester to focus only on the important parts that are shared across a given group of tickets. Our proposed method Absformer will be used to generate summaries for groups of tickets, and these summaries will be used as an additional source of information for the tester to solve a given ticket. The total number of tickets is 400k.  For the confidentiality of the ticket data, we cannot show samples of tickets and summaries. 

To sum up, the reviews from Amazon and Yelp are considered as \textbf{short documents}, whereas the tickets are considered as \textbf{long documents} with a significantly larger number of sentences that compose each ticket.


\subsection{Baselines}

We compare our proposed method Absformer against the following unsupervised abstravtive MDS baselines.

\subsubsection{Aspect + MTL \cite{coavoux}} An aspect-based clustering method is used to cluster reviews. Then, a generative LSTM-based model is trained using a multitask learning (MTL).

\subsubsection{MeanSum \cite{meansum}} The summarization process in Meansum starts by averaging the embeddings of documents that are obtained from the LSTM-based encoder in order to compute the embedding of the summary. Then, the resulting embedding is forwarded to an LSTM-based decoder to generate the summary. The training of the encoder and decoder is monitored using two loss functions namely the average summary similarity loss and the auto-encoder reconstruction loss.

\subsubsection{Copycat \cite{brazinskas_acl}} It is a VAE-based model that is trained using two types of loss functions: a reconstruction loss that recovers the original reviews from the latent representations, and a Kullback-Leibler (KL) divergence that penalizes the deviation of the computed posteriors from the priors. At the generation phase, the latent vectors of reviews are averaged to compute the summary representation that is decoded to generate the summary tokens.

\subsubsection{DenoiseSum \cite{amplayo_acl}} Many noisy versions of reviews are created using multiple noising techniques. LSTM is used in both the encoder to obtain contextualized embeddings, and the decoder to denoise reviews and generate summaries.

\subsubsection{RecurSum \cite{isonuma_ling}} A recursive Gaussian mixture models the sentence granularity in a latent space, and generates summaries with many granularities.

\subsection{Experimental Setup}

Our model is implemented using PyTorch, with NVIDIA RTX A6000. The dimension $h$ is equal to 768. For each dataset, the encoder is pretrained using the MLM for 50 epochs with a batch size of 32. The learning rate is 1e-4 with a weight decay 0.01, and a warmup phase of 80,000 steps. For fine-tuning of the encoder using the labeling information, the model is trained for 10 epochs with a batch size 32. The learning rate is 3e-5 with a linear decrease that starts with a warmup period in which the learning rate increases. The best model on the validation set is saved for phase II. 
 During the training step of phase II, we train the decoder for 40 epochs with a batch size of 8. The learning rate is 1e-4 with a weight decay 0.01, and a warmup phase of 20,000 steps. The maximum length of a sequence is 512, and we use token id 1 as the starting token of generated sequences. 
For summary decoding, the number of summaries $S$ is equal to 10. The top-K is 50 and top-p is 0.95.

 \begin{table*}
\centering
\begin{tabular}{@{}ccccc@{}}
\toprule
Model & \textbf{cosine$_{center}$} &\textbf{cosine$_{top-50}$}&\textbf{cosine$_{top-500}$} &\textbf{cosine$_{top-5000}$} \\ \midrule
Aspect + MTL \cite{coavoux}& 0.8126   & 0.7578    & 0.7252   & 0.7084  \\
 MeanSum \cite{meansum}& 0.8378   & 0.7723   & 0.7445  & 0.7273  \\
 Copycat \cite{brazinskas_acl}& 0.8541  & 0.7996   & 0.7688  & 0.7356  \\
 DenoiseSum \cite{amplayo_acl}& 0.8596   & 0.8011  & 0.7852  & 0.7624\\
 RecurSum \cite{isonuma_ling}& 0.8687    & 0.8135   & 0.7989  & 0.7783 \\
Absformer (w/o labeling information)   & \textbf{0.9034}   & \textbf{0.8553}  & \textbf{0.8453}  & \textbf{0.8133} \\
 \bottomrule
\end{tabular}
\vspace*{3mm}
\caption{Surrogate metrics on the ticket data.}
\label{ticket_results}
\end{table*}                       

\subsection{Experimental Results}

We evaluate the performance of our proposed method and baselines using three ROUGE \cite{rouge_score} scores: ROUGE-1 (R-1), ROUGE-2 (R-2), and ROUGE-L (R-L).

\subsubsection{Results on Amazon Dataset}

Table \ref{metrics}(a) shows the performance of different approaches on the Amazon reviews.  We show that our proposed method Absformer outperforms the baselines for R-1 and R-2. 
It is challenging to obtain a high R-L score because in our setting we summarize thousands of documents into a short summary, and it is unlikely to obtain a long sequence from the summary that exactly matches a sentence from ground truth summaries. 

In order to justify the importance of each component in our proposed method, we present the results of an ablation study for Absformer. In Absformer (w/o labeling information), we generate summaries for clusters of documents that are obtained from clustering the MLM-based embeddings using k-means. In Absformer (w/o decoder's initialization), the decoder is initialized from the vanilla DistilBERT instead of using the corresponding parameters from the encoder to initialize the embedding layer, the decoder block, and the language model head of the decoder. In Absformer (w/o pretraining), the encoder is fine-tuned directly on the target dataset. In Absformer (unweighted cross-entropy), all the weights $w_{d^{c(i)}_i}$ are equal to 1. Our full model Absformer outperforms all four system variations which supports the importance of (1) initializing the embedding layer, the decoder block, and the language model head of the decoder with the corresponding parameters from the encoder, (2) the labeling information, (3) the pretraining of the encoder, and (4) the unweighted cross-entropy loss for training the decoder $N$.

Compared to using BERT architecture in the encoder and decoder, the DistilBERT-based architecture of Absformer is \textbf{68\%} faster in the generation phase, with no significant drop in the ROUGE scores of the testing phase (retaining \textbf{98\%} of BERT performance in the summary generation).
\subsubsection{Results on Yelp Dataset}
Table \ref{metrics}(b) shows the performance of different approaches on the Yelp reviews collection. Consistent with Amazon dataset, our results on Yelp dataset shows that Absformer outperforms all the baselines and system variations for both R-1 and R-2.

\subsubsection{Results on Ticket Data from Network Equipment}
Unlike Amazon and Yelp datasets where there is a subset of ground truth summaries to report ROUGE scores in the testing phase, there are no ground truth summaries in the testing phase for the ticket data. Therefore, we cannot report ROUGE scores on this dataset to assess the quality of the generated summaries. Instead, surrogate metrics are used to evaluate the quality of the generated summaries for the ticket data. Similar to MeanSum \cite{meansum}, as a first surrogate metric, we report the average cosine similarity between the encoded generated summary and the embedding of a cluster for all $K$ clusters ($cosine_{center}$). The second surrogate metric is the average cosine similarity between the encoded generated summary and the top-$k$ tickets in each cluster for all $K$ clusters ($cosine_{top-k}$). The labeling information is not available in the ticket data, so Absformer (w/o labeling information) is used to generate summaries.

Table \ref{ticket_results} shows that Absformer achieves better surrogate metrics than the baselines. The embedding of the generated summary by Absformer is close to the center's embedding with an average cosine similarity around 0.9. For top-$k$ tickets in each cluster, when $k$ increases, the average cosine similarity between the encoded summary and the embeddings of the top-$k$ tickets decreases for both our method and baselines. This indicates the difficulty of summarizing a large number of documents in general, which simulates real case scenarios, and therefore is an important research direction. 

\begin{figure*}[ht!]
\centering
\includegraphics[width=0.80\textwidth]{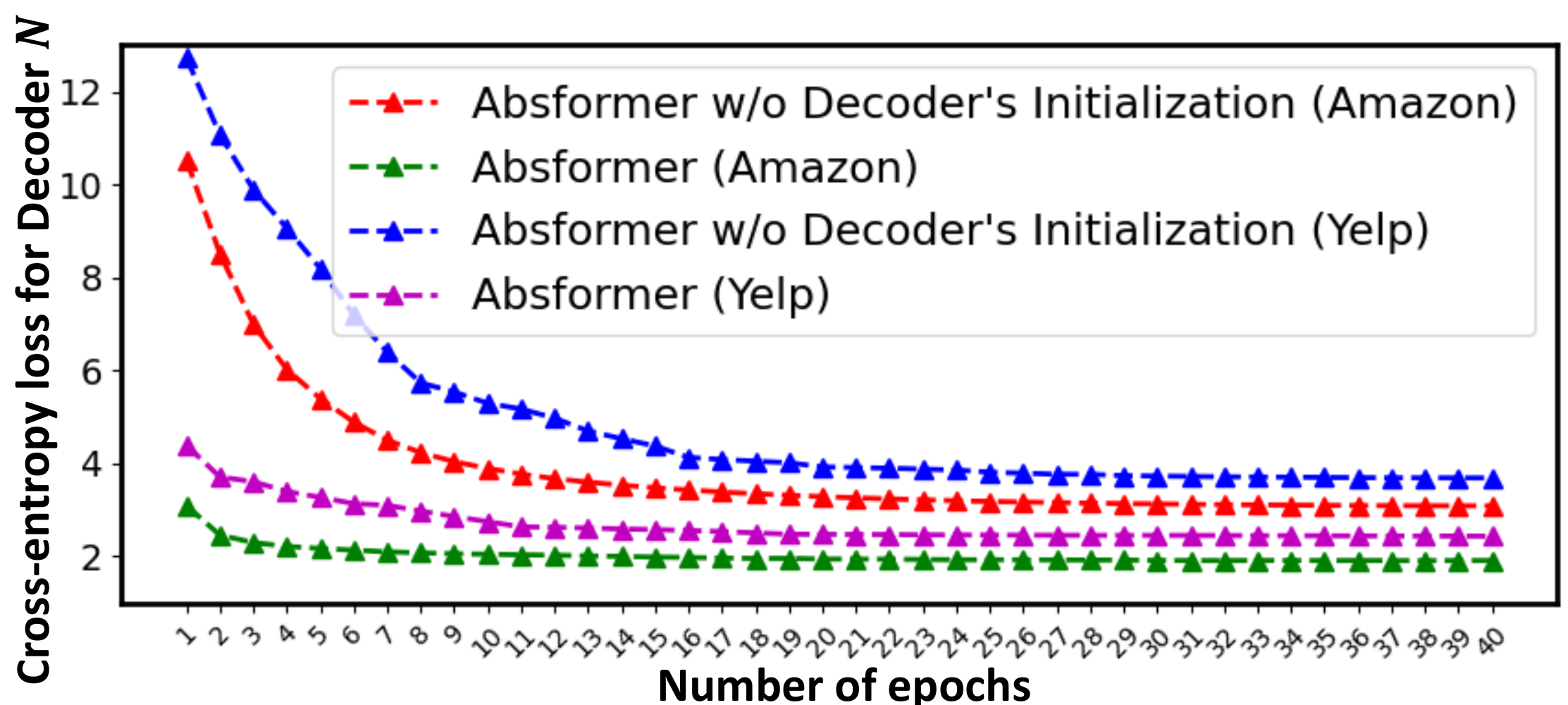}
\caption{Comparison of the cross-entropy loss of the decoder $N$ when the embedding layer, the decoder block, and the language model head of $N$ are not initialized using the encoder's parameters.  The red plot represents the cross-entropy of Absformer without decoder's initialization for Amazon, and is compared against the green plot that represents the cross-entropy of Absformer for Amazon; the blue plot represents the cross-entropy of Absformer without decoder's initialization for Yelp, and is compared against the magenta plot that represents the cross-entropy of Absformer for Yelp.
}
\label{visualization}
\end{figure*}
\subsubsection{Learning Curves of Absformer}
In Figure \ref{visualization}, we show the cross-entropy loss of the decoder $N$ on the validation set. For both datasets, initializing the embedding layer, the decoder block, and the language model head of $N$ with the corresponding parameters from the encoder $F$ leads to a lower cross-entropy loss and a faster convergence. As shown in Figure \ref{visualization}, our full model Absformer with the initialization applied to the decoder does not need to be trained for 40 epochs to converge. Training the decoder $N$ for only 10 epochs is sufficient to obtain a small cross-entropy loss on the validation set.

\subsubsection{Examples of Generated Summaries}
\begin{figure*}[ht!]
\centering
\includegraphics[width=0.80\textwidth]{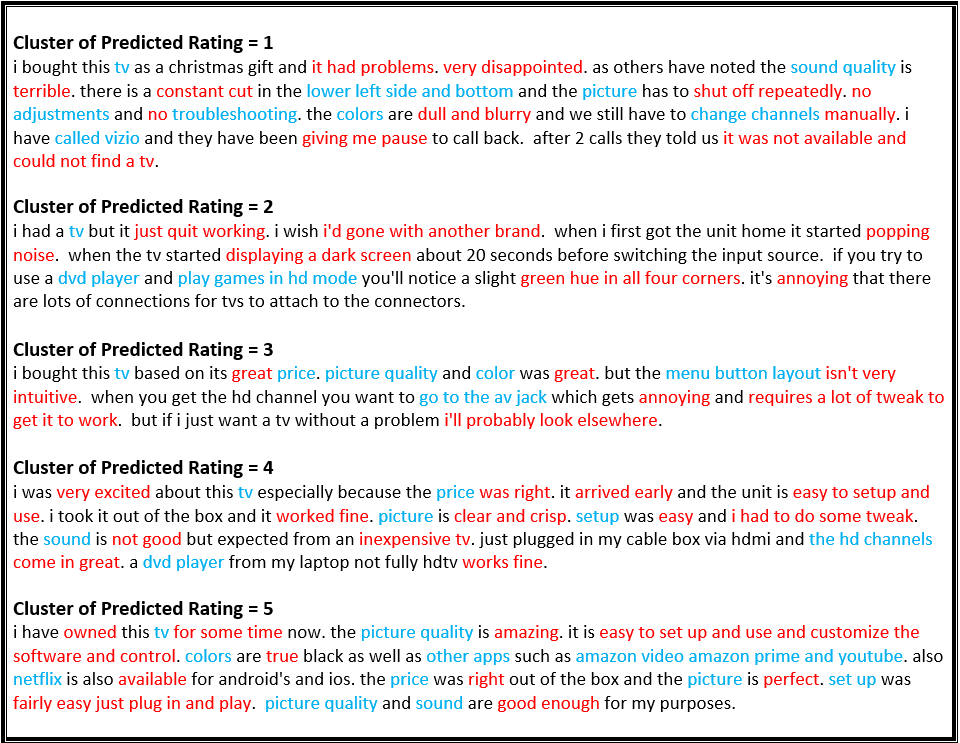}
\caption{Example of generated summaries for the TV product from Amazon. The light blue represents multiple aspects that are used to evaluate TVs. The red color represents either the description of aspects or the sentiment of the users.}
\label{summaries}
\end{figure*}

In Figure \ref{summaries}, we show examples of generated summaries from Absformer for the TV product from the Amazon dataset. We generate summaries for all 5 clusters, where each cluster represents a different predicted rating. Figure \ref{summaries} highlights multiple aspects that are captured by our summarizer. For example, the summaries cover the \textit{sound quality}, \textit{picture quality}, \textit{colors}, \textit{adjustments}, \textit{troubleshooting}, \textit{price}, \textit{menu button}, \textit{setup}, etc. The summary provides a description about the functionality of each aspect, where low ratings focus on problems and malfunctions, and high ratings focus on the fully functioning aspects. The summaries also highlight the sentiments of the users where negative sentiment is expressed in low-rating clusters using words such as \textit{disappointed}, and \textit{terrible}, and positive sentiment is expressed in high-rating clusters using words such as \textit{very excited}, and \textit{good enough}. Our summaries are fluent, and capture the main points of the reviews.

\section{Conclusions}
In this paper, we proposed a new unsupervised MDS method denoted by Absformer. We have shown that our Transformer-based two-phased model generates coherent and fluent summaries for clusters with semantically similar documents. Our Transformer-based decoder is trained by generating the documents from the embeddings that are obtained from the frozen encoder. Then, the trained decoder generates summaries from the centers of clusters. Improving the time and memory complexity, and achieving fast convergence are important points that are considered in our proposed model. We experimented on three datasets, and demonstrated that our new method outperforms the state-of-the-art baselines, and generalizes to
document collections from multiple domains. An ablation study shows the importance of four components: (1) initializing the embedding layer, the decoder block, and the language model head of the decoder with the corresponding parameters from the encoder, (2) the labeling information, (3) the pretraining of the encoder, and (4) the unweighted cross-entropy loss. 

Future work includes both (1) studying the possibility to incorporate Absformer into a variational autoencoder (VAE) framework for end-to-end training of both the encoder and decoder, and (2) summarizing reviews for each product to improve R-L score.

\bibliographystyle{ACM-Reference-Format}
\bibliography{main}
\appendix

\end{document}